\documentclass[letterpaper, 10 pt, journal, twoside]{IEEEtran}

\usepackage{times}

\usepackage{multicol}
\usepackage[bookmarks=true]{hyperref}
\usepackage{graphicx}
\usepackage{tabularx}
\usepackage{subfigure}
\usepackage{amsmath}
\usepackage{amssymb}
\usepackage{amsmath}
\usepackage{cite}
\usepackage[ruled,vlined]{algorithm2e}
\usepackage{multirow}
\usepackage{caption}
\usepackage{dcolumn}
\usepackage{physics}
\usepackage{hyperref}
\usepackage{color,soul}

\begin{document}

\title{Crash Landing onto ``you'': Untethered Soft Aerial Robots for Safe Environmental Interaction, Sensing, and Perching}

\author{Pham Huy Nguyen\\RSS Pioneers 2024 Research Statement\\Empa (Swiss Federal Laboratories for Materials Science and Technology) and Imperial College London \\ Email: huy.pham@empa.ch}
  
\maketitle
 
\IEEEpeerreviewmaketitle

There are various desired capabilities to create aerial forest-traversing robots capable of monitoring both biological and abiotic data. The features range from multi-functionality, robustness, and adaptability. These robots have to weather turbulent winds and various obstacles such as forest flora and wildlife thus amplifying the complexity of operating in such uncertain environments. The key for successful data collection is the flexibility to intermittently move from tree-to-tree, in order to perch at vantage locations for elongated time. This effort to perch not only reduces the disturbance caused by multi-rotor systems during data collection, but also allows the system to rest and recharge for longer outdoor missions. Current systems feature the addition of perching modules that increase the robots’ weight and reduce its overall endurance. Thus in our work, the key questions studied are:~\textbf{``How do we develop a single robot capable of metamorphosing its body for multi-modal flight and dynamic perching?''},~\textbf{``How do we detect and land on perchable objects robustly and dynamically?''}, and~\textbf{``What important spatial-temporal data is important for us to collect?''}

Spatial-temporal maps play a crucial role for researchers seeking to comprehend patterns and trends of regional biodiversity and abiotic environmental variables~\cite{altermatt2020uncovering}. They inform researchers about climate patterns, environmental conditions, and local population of animals and plants over space and time. In order to collect data for these maps, there is a need for sensors that monitor both the region’s inhabitant and habitat conditions simultaneously. This progress will enable researchers to develop more sophisticated conservation and climate strategies, enhancing efforts to protect biodiversity in these delicate habitats. However, to create these maps of the understory layer of forests has been an extremely difficult task, with multiple sensors needing to be placed and collected manually via human effort at various locations~\cite{hamaza2020sensor}. The introduction of robots for sensor placement have potential, but the technology is still at its nascent stage~\cite{hamaza2020sensor,aucone2023drone}. 

We have seen numerous bio-inspired principles as the bedrock of innovative solutions for aerial perching robots~\cite{chellapurath2023bioinspired,hang2019perching,lan2024aerial,zheng2024albero,zufferey2022ornithopters}. Similar to grasping and gripping principles, it has ranged from fibre-based dry adhesives like gecko-inspired solutions, to deployable microspines, tensile perching like spiders, and avian-inspired grippers~\cite{meng2022aerial}. For forest environments, dry adhesives often struggle because of the roughness of the tree branches and deployable microspines are often paired with another grasping mechanism to enhance gripping quality. Tensile perching and avian-inspired grippers, on-the-other hand, require additional modules plugged onto a commercial unmanned aerial vehicle, thus increasing the overall weight of the robot, reducing the possible payload it could carry, and therefore compromising mission endurance.

\begin{figure}[t!]
\centering
\includegraphics[width=0.47\textwidth]{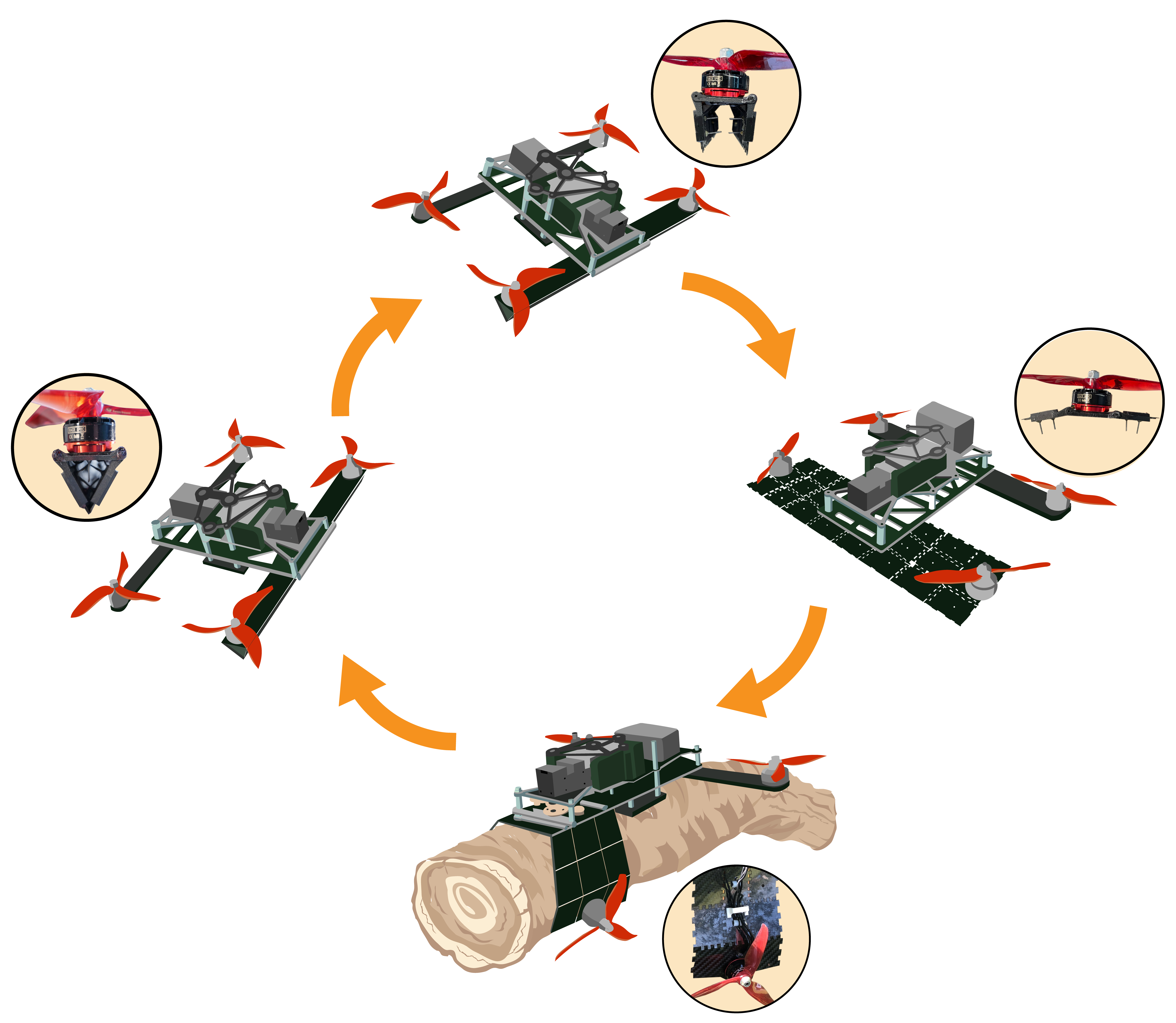}
\setlength{\belowcaptionskip}{-5pt}
\caption{\textbf{The Morphing Arm Aerial Robot} - Transitioning from flight to perching.}
\label{fig:fig1}
\vspace{-1.5em}
\end{figure}

As a result, current roboticists have started to explore shape-changing, reconfigurable, and metamorphic hardware that are very lightweight and/or emphasize shared functionality to adapt to the locomotion requirements in different environments, as seen in Fig.~\ref{fig:fig1}. Thus, aiming to reduce the additional weight and components required for each mode of mobility. Furthermore, the addition of embodied compliance in aerial robot bodies will improve their adaptability to the environmental uncertainties and physical disturbances in these forest environments. 

With the research objective centralized around the development of a metamorphic robot body and its interaction with different environments, one exciting strategy to develop aerial robots is through materials-based approaches seen in soft robotics~\cite{miriyev2020skills,nguyen2022adopting}. Soft robotic methodologies open pathways to develop aerial perching robots that are imbued with soft robotic components, such as soft structures, sensing, and actuation~\cite{stella9science,nguyen2022adopting}. The goal would not be to create a completely soft-bodied aerial robot, because classically soft systems still suffer from slow actuation speeds, but use a combination of flexible, rigid and soft materials with encoded mechanical shape-changing and sensing capabilities instead. \textbf{I believe this design methodology will lead to aerial robot designs that can manipulate and sense its shape for shared functionality, but are still safe and robust enough to interact and transition between flight and perching.} 

\section{Past Research Experience}

Previously, through the study and characterization of intrinsically soft materials, with a focus on hyperelastic elastomers and multi-layered fabric materials, my team and I were able to create a computational model library that captured both their anisotropic and isotropic behavior~\cite{nguyen2020design}. This library was then utilized for optimizing the actuator designs for payload capacity and motion capabilities. These models were extended to investigate the capabilities of soft wearable assistive/rehabilitative devices, graspers, and continuum arms, that were also manufactured and tested. Further, we also embedded distributed sensors made of conductive rubber stretch sensors and fabrics to track the movement of soft continuum units that could perform multi-axis bending, helical twisting, and contraction~\cite{nguyen2020design,nguyen2019soft,nguyen2019fabric}. With proprioceptive and motion capture sensors, we also proposed control models to improve the system tracking performance of these units~\cite{qiao2022model,nguyen2020towards}.

With the goals centered around the development of physically intelligent aerial robot platforms, our team has worked on aerial robots that are highly adaptable and physically robust to collisions with the environment~\cite{nguyen2022adopting}. We developed an aerial robot with a fully soft body made of inflatable fabrics co-developed with a fabric-based bistable grasper~\cite{nguyen2022_soBAR}. This work emphasized the potential of utilizing a variable stiffness soft body for mitigating impacts due to collisions, landing, and dynamic vertical perching. The conclusion of this work was the improved robustness of a soft-bodied frame in comparison to conventional frames and the frame’s ability to absorb impact when dynamically perching to extend contact time with the grasping object, for highly successful aerial grasping. 

Further, our collaborative efforts have seen aerial-aquatic drones donned with a soft remora-inspired suction cup, capable of perching itself on uneven structures underwater and on land. Thus extending the importance of perching for aerial robots in multiple mediums~\cite{li2022aerial}.
 
More recently, we explored the metamorphic design methodology to establish shared functionality in the aerial robot arms, utilizing the same robot hardware~\cite{zheng2023metamorphic}. In essence, our aerial robot arms were capable of switching between flight and grasping for perching, without the need of an external gripper. Because of its origami-based design, the robot was able to wrap around perching structures of different sizes. This shared utility of the arms was able to reduce the overall weight of the system, which contributed to operational endurance as well.\textbf{ These soft robotic design methodologies showed enhanced flexibility in terms of aerial metamorphic robot development, along with capabilities to interact well with the environment with added robustness. }

\section{Future Work}
Recently, different versions of morphing arm aerial robots (MAARs) have been developed, showcasing their ability in enabling adjustable flight dynamics and facilitating horizontal or vertical perching~\cite{jia2023aerial,ruiz2022sophie,zheng2023metamorphic}. Although these works establish an encouraging benchmark, I aim to extend the capabilities of these systems towards successful field missions by:
\begin{enumerate}
    \item Developing MAAR systems that fully explore the complexities of high-speed, high-impact dynamic perching from various angles on more complex objects. 
    \item Tackling the complexities of graspable branch detection utilizing onboard vision-based perception methods.  
    \item Equipping robot with sensors that are capable of collecting environmental data (such as acoustic or eDNA sensors) that can actually support ecologists and perform field missions alongside them.
\end{enumerate}    
\textbf{Dynamic Perching}: To enable high-impact dynamic perching, we would need the next version of our robot to absorb impact even better. This means taking a page from nature and enhancing our robot arms with elastic underactuated tendons~\cite{roderick2021bird}. This would aid in absorbing impact energy at the joints, during the free-fall dynamic perching phase, while still providing softness to adapt to the perching object and surface. A similar development process will be utilized to make the body structure of the drone for impact absorption without hindering perching. 

To perform autonomous dynamic perching would require a post-stall maneuver performed by the aerial robot, by adjusting the flight speed, flight trajectory, perching mechanism, and motor disarming time. This needs to be a harmonious process between perception, planning and control. The same principles are relevant to dynamic takeoff scenarios as well, where the arms would need to swiftly transition back to its rigid configuration as the motor and propeller pairs turn on.  

\textbf{Vision-based Branch Detection}: To detect perch-able tree branches with an onboard sensor is a considerable challenge due to the randomness of orientation, size, shape, overlapping nature, and occlusive textures surrounding them, such as leaves~\cite{livny2010automatic, li2017rapid}. With a vision-based sensor and processing power of an onboard computer, it complicates the problem further. Past work has seen convolution neural networks identify and segment trees, branches, twigs, and leaves but these datasets remain limited~\cite{digumarti2019approach,cordts2016cityscapes,lin2014microsoft}. The best approach for our team would be through a synthetic dataset generation pipeline instead to ease the difficulties of data collection and labeling, as well as enhances the amount of feature variability possible.  

\textbf{Environmental Sensing Network:}  
The eventual goal of creating a wireless network of robots monitoring a forest area can be performed with a single robot intermittently going from location-to-location and eventually scaling toward multiple robots. Onboard each robot, we aim to have a camera, eDNA, and acoustic sensors to track biodiversity and possibly poachers, as well as wide-spectrum sensors for measuring gas level, temperature, humidity, air pressure, $\mathrm{CO_2}$, etc.

\bibliographystyle{ieeetr}
\bibliography{references}

\end{document}